\documentclass[runningheads]{llncs}
\usepackage{graphicx}
\usepackage{subcaption}
\usepackage{multirow}
\begin{document}
\title{L3Cube-MahaNLP: Marathi Natural Language Processing Datasets, Models, and Library\thanks{Supported by L3Cube Pune}}
\titlerunning{L3Cube-MahaNLP: Marathi Natural Language Processing}
%
\author{
Raviraj Joshi
}
\authorrunning{R. Joshi et al.}
%
\institute{
Indian Institute of Technology Madras, Chennai, Tamilnadu India \and
L3Cube Pune \\
\email{ravirajoshi@gmail.com}}
\maketitle              
\begin{abstract}
Despite being the third most popular language in India, the Marathi language lacks useful NLP resources. Moreover, popular NLP libraries do not have support for the Marathi language. With L3Cube-MahaNLP, we aim to build resources and a library for Marathi natural language processing. We present datasets and models for supervised tasks like sentiment analysis, named entity recognition, and hate speech detection. We have also published a monolingual Marathi corpus for unsupervised language modeling tasks. Overall we present MahaCorpus, MahaSent, MahaNER, and MahaHate datasets and their corresponding MahaBERT models fine-tuned on these datasets. We aim to move ahead of benchmark datasets and prepare useful resources for Marathi. The resources are available at  https://github.com/l3cube-pune/MarathiNLP.\\

\end{abstract}

\section{Introduction}

Text-based natural language processing (NLP) has become mainstream with libraries like NLTK and Spacy \cite{loper2002nltk,honnibal2017natural}. These libraries support basic rule-based features like tokenization, stemming, lemmatization, etc., and more complex machine learning-based features like classification, named entity recognition, parts of speech (POS) tagging, etc. The Transformers library \cite{wolf2020transformers} provides APIs to support state-of-the-art PyTorch or TensorFlow models for advanced tasks like summarization, machine translation, question answering, sentence similarity, etc \cite{qiu2020pre}. While production-ready libraries and models have been available for English same is not true for low resource languages. In this work, we specifically focus on the Indian low resource language Marathi. 

Marathi is the third most popular language in India spoken by around 83 million people \cite{joshi2022l3cube}. The language is native to the state of Maharashtra and is the most spoken language after Hindi and Bengali \cite{joshi2019deep}\cite{alam2020bangla}. Despite Maharashtra being the educational and industrial hub of India, Marathi NLP has not received enough attention from academia or industry. We found that Marathi lacks behind in the most basic form of NLP resources like monolingual text corpus. With L3Cube-MahaNLP we aim to build more useful resources for the Marathi language. Our vision is to make Marathi a resource-rich language and develop a dedicated library with focused contributions. Although recently resources have been Indic languages still Marathi stands behind languages like Hindi and Bengali. We aim to make dedicated contributions to the low-resource Marathi language in terms of datasets and models for various NLP tasks. Another objective of this work is to move away from small benchmark datasets which may help advance the state of the art but are not normally useful in production. This is the first work to build production-quality datasets and models for the Marathi language.      
All the resources are publicly shared on github\footnote{https://github.com/l3cube-pune/MarathiNLP}.

\section{Features}
Currently, L3Cube-MahaNLP supports tasks like tokenization, word vectors, monolingual BERT models, sentiment analysis, named entity recognition, next token prediction, and hate speech detection. We release datasets and models for these tasks on github and model repository respectively. We are in the process of wrapping these models under python API's that can be consumed directly. As of now, the models can be accessed using the hugging face API's. The different tasks, datasets, and models are described below.
\begin{itemize}
    \item \textbf{MahaCorpus} \cite{joshi2022l3cube}\footnote{https://github.com/l3cube-pune/MarathiNLP}: 
    MahaCorpus is a Marathi monolingual corpus with 24.8M sentences and 289M tokens. Combined with other publicly available resources we have a total of 752M tokens. The dataset has been scraped from the internet using news and non-news sources. The major chunk of the dataset comes from news sources with 212M tokens and the remaining 76.4M tokens come from non-news literature work. Both the sources have been released separately to enable further research in an individual area. The monolingual data set can be used to train models using unsupervised language modeling tasks. 
    \item \textbf{MahaBERT} \cite{joshi2022l3cube}\footnote{https://huggingface.co/l3cube-pune/marathi-bert}\footnote{https://huggingface.co/l3cube-pune/marathi-roberta}\footnote{https://huggingface.co/l3cube-pune/marathi-albert}\footnote{https://huggingface.co/l3cube-pune/marathi-albert-v2}: 
    The BERT represents a deep bi-directional Transformer based model trained using a large unlabelled corpus. These pre-trained models have been shown to produce state-of-the-art results on a variety of downstream tasks. There are different variations of BERT models like base-BERT, AlBERT and RoBERTa which are also considered in this work. From the multilingual perspective, there are three main models which can also be used with the Marathi language. These include multilingual-BERT \cite{devlin2018bert},  XLM-R based on RoBERTa \cite{conneau2019unsupervised}, and IndicBERT \cite{kakwani2020inlpsuite} based on AlBERT. The L3Cube-MahaCorpus along with other publicly available Marathi corpus is used to train these three variations of BERT using MLM objective. These variations are based on base-BERT, AlBERT, and RoBERTa architecture and are termed as MahaBERT, MahaAlBERT, and MahaRoBERTa respectively. These monolingual models perform better than their multi-lingual counterparts on Marathi downstream tasks.
    \item \textbf{MahaGPT} \cite{joshi2022l3cube}\footnote{https://huggingface.co/l3cube-pune/marathi-gpt}:
    GPT2 is a generative transformer model trained using causal language modeling (CLM) objective \cite{radford2019language}. It is also a class of self-supervised models trained to predict the next work on the unsupervised data. MahaGPT is GPT2 model trained on full Marathi Corpus. The model can be used for next word prediction tasks or auto-completion tasks. The model can also be used to generate full sentences from the initial text.
    \item \textbf{MahaFT} \cite{joshi2022l3cube}\footnote{https://github.com/l3cube-pune/MarathiNLP}: 
    Pre-trained word embeddings are commonly used to initialize the embedding layer of the neural networks. These can also be used for similarity-based tasks. These distributed representations are trained on large unlabeled corpus and are useful for many downstream tasks. The FastText word embeddings are popular for morphologically rich languages \cite{bojanowski2017enriching}. It represents the word as a bag of character n-grams thus avoiding any out of vocabulary word. We release MahaFT pre-trained fast text word vectors on full Marathi corpus. The word embeddings are shown to work better than the existing word vectors available for Marathi.
    \item \textbf{MahaSent} \cite{kulkarni2021l3cubemahasent}\footnote{https://huggingface.co/l3cube-pune/MarathiSentiment}: 
    MahaSent is the first major Marathi Sentiment Analysis Dataset. The sentiment analysis task consists of Marathi tweets categorized as positive, negative, and neutral. The dataset consists of 12114 train, 2250 test, and 1500 validation examples. We also release the BERT model fine-tuned on this dataset that can be directly used to predict sentiment labels. The models released support two-way classification (positive and negative) and three-way classification (positive, negative, and neutral).
    \item \textbf{MahaNER} \cite{patil2022l3cube}\footnote{https://huggingface.co/l3cube-pune/marathi-ner}:
    MahaNER is the first major gold standard named entity recognition dataset in Marathi. It contains 25,000 manually tagged sentences categorized according to the eight entity classes. These entities annotated in the dataset include names of locations, organizations, people, and numeric quantities like time, measure, and other entities like dates and designations. The dataset is released in IOB and non-IOB notations. The dataset is divided into 21500, 2000, and 1500 train, test, and validation samples respectively. The BERT model for this token classification task is released publicly on the model hub. 
    \item \textbf{MahaHate} \cite{velankar2022l3cube}\footnote{https://huggingface.co/l3cube-pune/mahahate-bert}\footnote{https://huggingface.co/l3cube-pune/mahahate-multi-roberta}: 
    MahaHate is the first major Hate Speech Dataset in Marathi. The dataset is curated from Twitter and annotated manually. Our dataset consists of over 25000 distinct tweets labeled into four major classes i.e hate, offensive, profane, and not. The dataset is divided into 21500, 2000, and 1500 train, test, and validation samples respectively. The BERT models trained on this dataset are released publicly and can be directly utilized to perform two-way classification (hate and non-hate) and four-way classification (hate, offensive, profane, and neutral). With the rise of hateful content on social media platforms, the availability of such models becomes more important for regional languages.
    
\end{itemize}

\section{Impact}

The monolingual Marathi models released in this work have been shown to work better than the currently available alternatives. A study conducted for Marathi named entity recognition highlights the importance of our models and datasets \cite{litake2022mono}. A similar study for Marathi text classification and specifically hate speech detection was conducted in \cite{velankar2022mono}. The MahaBERT models have been shown to work well on sentence classification and token classification tasks. Some of the other works done as a part of this project include \cite{kulkarni2021experimental}\cite{velankar2021hate}.

\section{Conclusion}

We present L3Cube-MahaNLP, a host of Marathi datasets, models, and library. We highlight the lack of resources for the Marathi language and built some most basic resources. The different datasets built as a part of this work include L3Cube-MahaCorpus, MahaSent, MahaNER, and MahaHate. We also release transformer models trained on these datasets on the model hub.

In the future, we want to further expand on datasets, and models and focus more on Marathi natural language generation tasks. We will wrap all the models under a pip package which can currently be accessed using hugging face API. 

\section*{Acknowledgements} 
Multiple L3Cube-Pune student groups have contributed to this work. We thank all the students for their dedicated contributions.

\bibliographystyle{splncs04}
\bibliography{main}
\end{document}